\documentclass{article}

\usepackage[preprint]{neurips_2022}
\usepackage[utf8]{inputenc} 
\usepackage[T1]{fontenc}    
\usepackage{hyperref}       
\usepackage{url}            
\usepackage{booktabs}       
\usepackage{amsfonts}       
\usepackage{nicefrac}       
\usepackage{microtype}      
\usepackage{xcolor}         
\usepackage{graphicx}
\usepackage{amsmath}
\usepackage{amsfonts}
\usepackage{multirow}
\usepackage{amsfonts}
\usepackage{tikz}
\usetikzlibrary{calc}
\usepackage{tikz-qtree}
\usetikzlibrary{trees}
\usepackage[edges]{forest}
\usetikzlibrary{shadows,arrows.meta}
\usepackage{makecell}
\usepackage{color}
\usepackage{CJKutf8}
\usepackage{rotating}

\usepackage{siunitx}
\usepackage{colortbl}
\usepackage{makecell}
\usepackage[ruled,linesnumbered]{algorithm2e}
\usepackage{algpseudocode} 
\usepackage{subcaption}
\usepackage{bbm}
\definecolor{maroon}{cmyk}{0,0.87,0.68,0.32}

\usepackage[utf8]{inputenc} 
\usepackage[T1]{fontenc}    
\usepackage{hyperref}       
\usepackage{url}            
\usepackage{booktabs}       
\usepackage{amsfonts}       
\usepackage{nicefrac}       
\usepackage{microtype}      
\usepackage{xcolor}         

\title{UPainting: Unified Text-to-Image Diffusion Generation with Cross-modal Guidance}

\author{%
  Wei Li, Xue Xu, Xinyan Xiao\thanks{Corresponding author.}, Jiachen Liu, Hu Yang, Guohao Li\\
  \textbf{Zhanpeng Wang, Zhifan Feng, Qiaoqiao She, Yajuan Lyu, Hua Wu}\\
  Baidu Inc.\\
  \{\texttt{liwei85,xiaoxinyan,sheqiaoqiao,lvyajuan}\}@baidu.com \\
}

\begin{document}

\maketitle

\begin{figure}[!h]
    \centering
    \setlength{\leftskip}{-40pt}
    \includegraphics[width=6.5in]{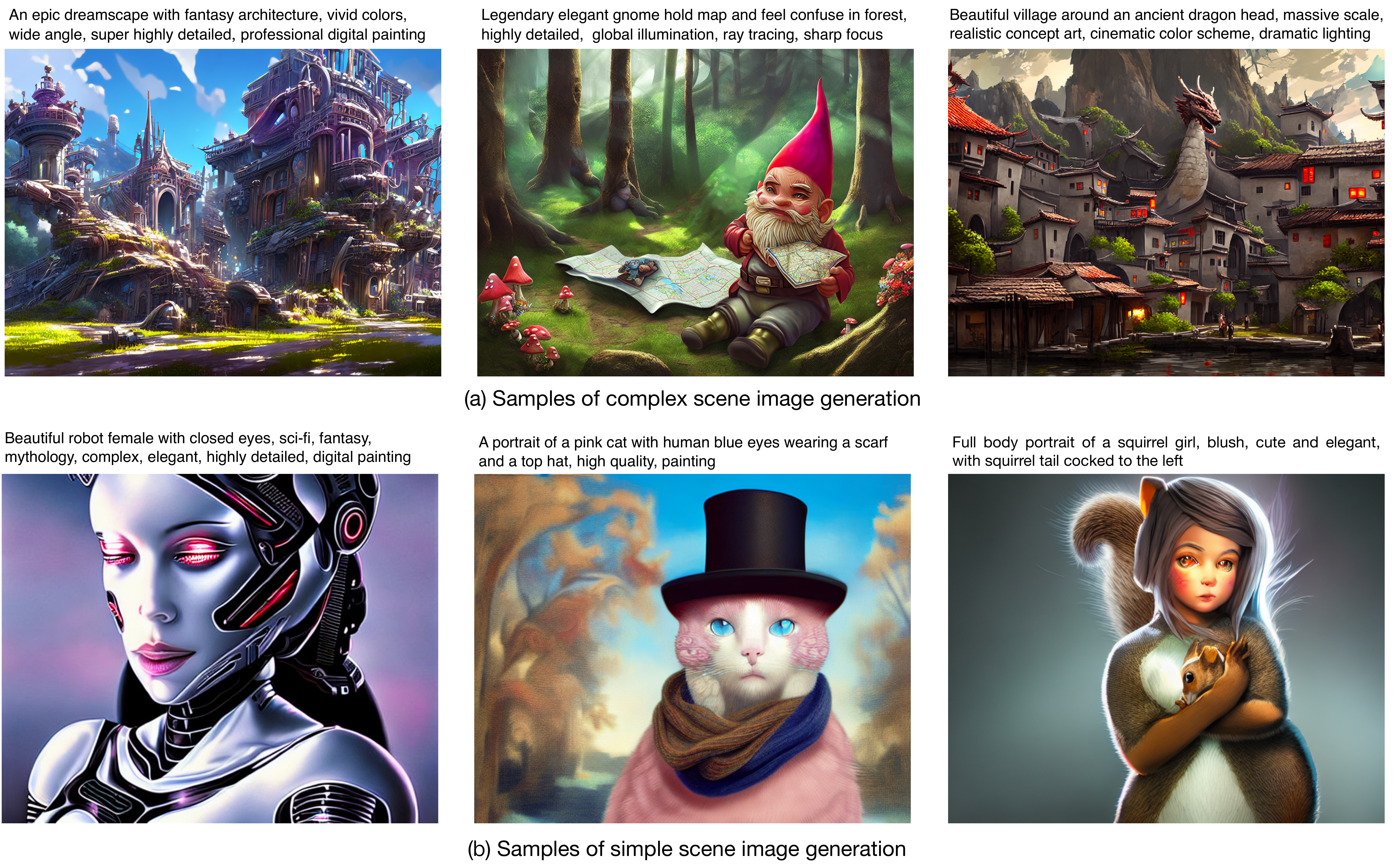}
    \caption{Selected samples of both complex-scene and simple-scene images generated by UPainting.}
    \label{fig:leading_samples}
\end{figure}

\begin{abstract}
Diffusion generative models have recently greatly improved the power of text-conditioned image generation.
Existing image generation models mainly include text conditional diffusion model and cross-modal guided diffusion model, which are good at small scene image generation and complex scene image generation respectively.
In this work, we propose a simple yet effective approach, namely UPainting, to unify simple and complex scene image generation, as shown in Figure~\ref{fig:leading_samples}.
Based on architecture improvements and diverse guidance schedules, UPainting effectively integrates cross-modal guidance from a pretrained image-text matching model into a text conditional diffusion model that utilizes a pretrained Transformer language model as the text encoder.
Our key findings is that combining the power of large-scale Transformer language model in understanding language and image-text matching model in capturing cross-modal semantics and style, is effective to improve sample fidelity and image-text alignment of image generation.
In this way, UPainting has a more general image generation capability, which can generate images of both simple and complex scenes more effectively.
To comprehensively compare text-to-image models, we further create a more general benchmark, UniBench, with well-written Chinese and English prompts in both simple and complex scenes.
We compare UPainting with recent models and find that UPainting greatly outperforms other models in terms of caption similarity and image fidelity in both simple and complex scenes\footnote{https://upainting.github.io/}.

\end{abstract}



\section{Introduction}

Multimodal learning has made great strides by scaling models on large datasets of image-caption pairs collected from the Internet, where image-text cross-modal matching~\citep{radford2021learning,li2021align,li2020unimo,li2022unimo,yu2021ernie} and text-to-image generation~\citep{ramesh2021zero,rombach2022high,saharia2022photorealistic,ramesh2022hierarchical,nichol2021glide} are at the forefront.
Text conditional image generation includes simple scene image generation that usually depicts a specific object, and complex scene image generation that depicts an abstract or large scene with multiple objects, as shown in Figure~\ref{fig:leading_samples}.
Existing text-to-image diffusion models mainly focus on simple scene image generation, such as GLIDE~\citep{nichol2021glide}, DALL-E 2~\citep{ramesh2022hierarchical}, and Imagen~\citep{saharia2022photorealistic}.
They mainly utilize text encoders trained on textual data or paired image-text data to capture the semantics and compositionality of natural language inputs.
However, plain text encoders are difficult to capture all aspects of complex text prompts, such as sophisticated art styles (e.g. fantasy art, cinematic composition, etc.), aesthetic features (e.g. super wide angle, bright atmosphere, dramatic lighting, etc.) and abstract descriptions (e.g. epic dreamscape, time and pressure, etc.).


Pre-trained cross-modal matching models are effective in image-text alignment, and they typically capture rich cross-modal semantics and styles from large-scale corpora.
In this work, to unify simple and complex scene image generation, we propose a simple but effective approach UPainting, which enhances a text-conditional diffusion model with cross-modal guidance.
Specifically, a pre-trained transformer language model encodes the textual prompts as semantic conditions for the diffusion model, and a pre-trained image-text matching model simultaneously guides the diffusion process through cross-modal matching based on several crucial architecture improvements and diverse guidance schedules.
Our key findings is that combining a transformer language model and cross-modal matching models is very effective for improving sample fidelity, aesthetics and image-text alignment for diffusion image generation.
In this way, UPainting has a general image generation capability, which can generate both simple and complex scene images more effectively.
Some previous work have also explored to utilize pre-trained image-text matching models (i.e. CLIP~\citep{radford2021learning}) to guide unconditional or class-conditional diffusion models~\citep{Crowson2021dd256,Crowson2021dd512,liu2021more}.
Without powerful text encoders to encode language prompts, these models are usually difficult to generate hyper-realistic images or complex scenes with details.
Combining a large-scale transformer language model and cross-modal matching model is crucial for general text-conditional image generation.
Some comparison examples are shown in Figure~\ref{fig:comparison_small_scene} and Figure~\ref{fig:comparison_complex_scene}.

Since existing evaluation benchmarks mainly focus on simple scene image generation, such as the COCO dataset~\citep{lin2014microsoft} and the DrawBench~\citep{saharia2022photorealistic} dataset, we further propose a more comprehensive and challenging benchmark, UniBench, to evaluate general text-to-image generation ability. UniBench contains 200 well-written queries for both simple scenes and complex scenes in both Chinese and English. We compare UPainting with several recent models, including Stable Diffusion~\citep{rombach2022high} and Disco Diffusion~\citep{Crowson2021dd512}, and find that UPainting greatly outperforms other methods in terms of image-text alignment and image fidelity in both simple and complex scenes. We also evaluate UPainting on the COCO dataset, and it achieves much better performance (with 8.34 FID score) than previous work such as Stable Diffusion (at 14.24). The results further demonstrate the general capability of image generation by effectively combining the transformer language model and cross-modal matching models.


Key contributions of the paper include:
\begin{itemize}
    \item For the first time, we systematically study the problem of text-conditional image generation for both simple and complex scenes, and propose a simple yet effective method to unify image generation for simple and complex scenes.
    \item We discover that by effectively combining cross-modal matching models with pre-trained transformer language models can greatly improve sample fidelity and image-text alignment for diffusion image generation. This gives the model a general ability to generate images for both simple and complex scenes.
    \item We introduce UniBench, a comprehensive and challenging evaluation benchmark for both simple and complex scene image generation. On UniBench, our proposed method outperforms other work in both image-text alignment and image fidelity. 
\end{itemize}

\section{Method}
\label{headings}

\begin{figure}
    \centering
    \setlength{\leftskip}{-40pt}
    \includegraphics[width=6.5in]{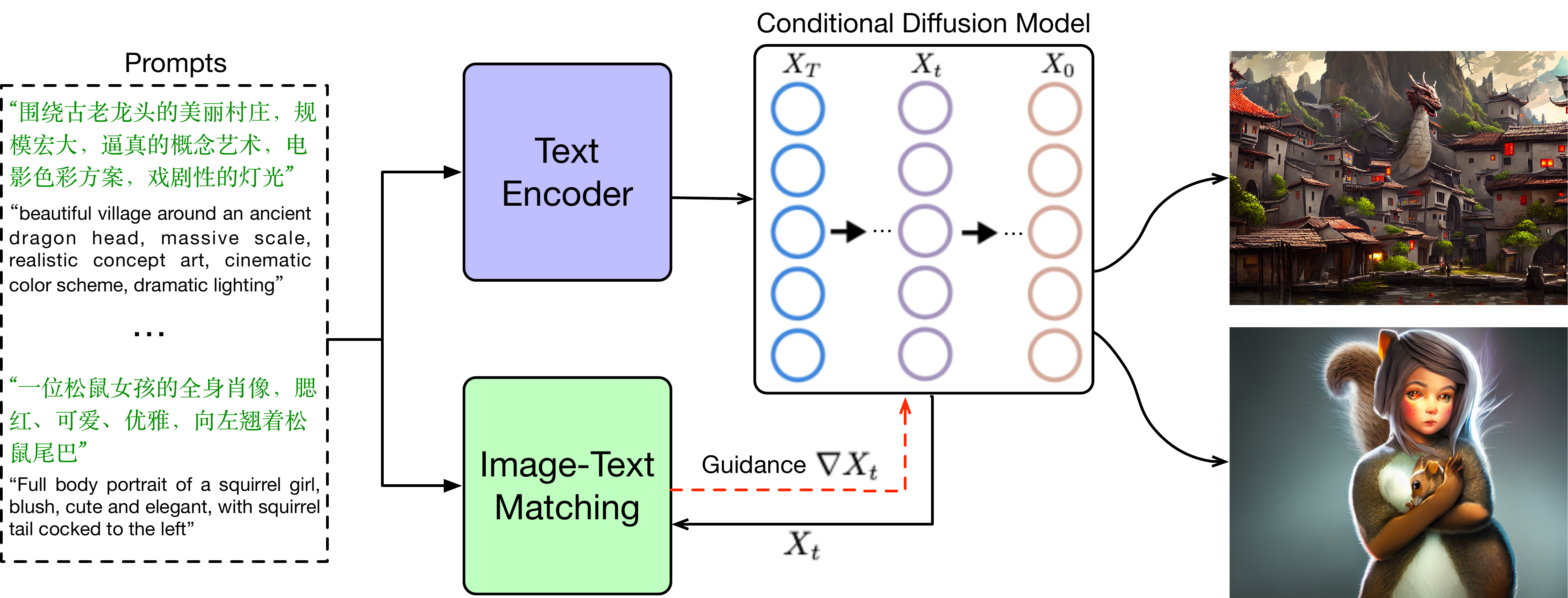}
    \caption{Illustration of the architecture of UPainting, which incorporates cross-modal matching models with a text-conditional diffusion model. UPainting has a general image generation capability, which can effectively generate simple scene images as well as complex scene images.}
    \label{fig:unimo_UPainting}
\end{figure}

UPainting consists of a text encoder, an image-text matching component, and a conditional diffusion model, as shown in Figure~\ref{fig:unimo_UPainting}.
The text encoder maps text prompts to a sequence of embeddings, and then the conditional diffusion model maps these embeddings to corresponding images.
During the diffusion process, the image-text matching component further guides each diffusion step by cross-modal matching, in order to better capture the complex cross-modal semantics and styles.
In the following subsections, we describe these components in detail.

\subsection{Text-conditional Diffusion Model}
Our training dataset consists of pairs $(x,y)$ of images $x$ and their corresponding captions $y$. Given a text prompt $y$, the text encoder encodes it into a sequence of embeddings $c$. Then the conditional diffusion model $P(x|c)$ learns to produce images $x$ conditioned on text embeddings $c$.

\paragraph{Text Encoder.} 
Creative text-to-image generation needs powerful semantic text encoders to capture the complexity and compositionality of arbitrary natural language text inputs.
Transformer-based language models (e.g., BERT\citep{devlin2018bert}, GPT\citep{brown2020language}, T5\citep{raffel2020exploring}, ERNIE\citep{sun2020ernie}) have led to leaps in textual understanding capabilities.
Imagen\citep{saharia2022photorealistic} has shown the effectiveness of large-scale pre-trained transformer language models to encode text for text-to-image generation.
Thus, we apply a pre-trained transformer language model as our text encoder and further update it during the training process of the diffusion model on large volumes of paired image-text data.


\begin{figure}
    \centering
    \setlength{\leftskip}{-40pt}
    \includegraphics[width=7in]{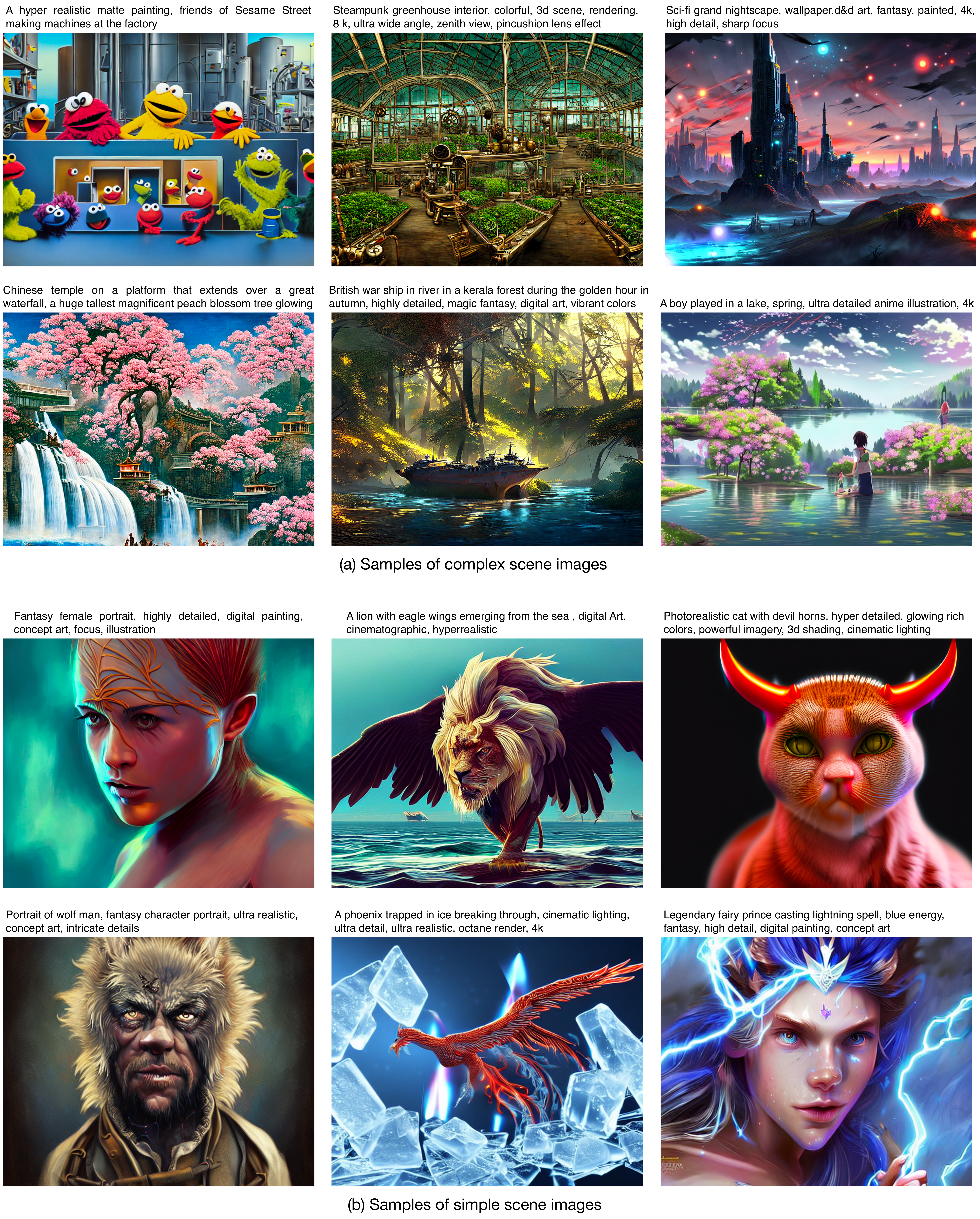}
    \caption{More samples of both complex-scene and simple-scene images generated by UPainting.}
    \label{fig:more_samples}
\end{figure}

\paragraph{Conditional Diffusion Model.}
Diffusion models are a class of generative models that convert Gaussian noise into samples from a learned data distribution via an iterative denoising sampling process.
In particular, sampling starts with Gaussian noise $x_T \sim \mathcal{N}(0, I)$ and produces gradually less-noisy samples $x_{T-1}$, $x_{T-2}$, $\dots$, until reaching a final sample $x_0$. 
Each timestep $t$ corresponds to a certain noise level, and $x_t$ can be thought of as a mixture of a signal $x_0$ with some noise $\epsilon$ where the signal to noise ratio is determined by the timestep $t$.
A diffusion model learns to produce a slightly more ``denoised'' $x_{t-1}$ from $x_t$. \citet{ho2020denoising} parameterize this model as a function $\epsilon_{\theta}(x_t, t, c)$ which predicts the noise component of a noisy sample $x_t$. To train these models, each sample in a minibatch is produced by randomly drawing a data sample $x_0 \sim q(x_{0})$, a timestep $t \sim \mathcal{U}(1, T)$, and noise $\epsilon \sim \mathcal{N}(0, I)$, which together give rise to a noised sample $x_t = \sqrt{\bar{\alpha_t}} x_{0} + \epsilon \sqrt{1 - \bar{\alpha_t}}$ (\ref{eq_xt}).
The training objective is a simple mean-squared error loss between the true noise and the predicted noise of the form:
\begin{equation}
    \mathbb{E}_{x_{0}, t, \epsilon, c} [\Vert \epsilon_{\theta}(x_t, t, c) - \epsilon \Vert^{2}]
    \label{eq_loss}
\end{equation}

To make the diffusion process conditional on the text prompts, we augment the underlying UNet~\citep{ho2020denoising} backbone of the diffusion model with the cross-attention mechanism.
The token sequence embeddings $c$ is mapped to the intermediate layers of the UNet via a cross-attention layer implementing $Attention(Q,K,V) = softmax(\frac{QK^{T}}{\sqrt{d}}) \cdot V$, where the intermediate representations of the UNet acting as the query $Q$, and the token sequence embeddings $c$ acting as the key $K$ and value $V$.

Classifier-free guidance~\citep{ho2022classifier} is a widely used technique to improve sample quality while reducing diversity in conditional diffusion models, which jointly trains a single diffusion model on conditional and unconditional objectives via randomly dropping $c$ during training (e.g. with $10\%$ probability). 
During sampling, the output of the model is extrapolated further in the direction of $\epsilon_{\theta} (x_t|c)$ and away from $\epsilon_{\theta}(x_t|\emptyset)$ as follows:
\begin{equation}
    \hat{\epsilon}_{\theta} (x_t | c) = \epsilon_{\theta}(x_t|\emptyset) + s \cdot (\epsilon_{\theta} (x_t|c) - \epsilon_{\theta}(x_t|\emptyset))
\end{equation}
\noindent where $s\ge1$ is the guidance scale. Larger $s$ will increase the fidelity of samples while reducing diversity. We set $s=8.0$ empirically in our experiments.

\subsection{Image-Text Matching Guidance}
\label{subs:img2txt_matching}

Dhariwal and Nichol~\citep{dhariwal2021diffusion} have explored to utilize a classifier $p(y|x)$ to improve a diffusion generator.
In particular, they train a classifier $p_{\phi}(y|x_t, t)$ on noisy images $x_t$, and then use gradients $\nabla_{x_t} log\ p_{\phi}(y|x_t, t)$ to guide the diffusion sampling process towards an arbitrary class label $y$.
To apply the same idea to text-to-image diffusion models, \citet{nichol2021glide} further replace the classifier with a CLIP~\citep{radford2021learning} model in classifier guidance.
A CLIP model consists of two separate pieces: an image encoder $f(x)$ and a caption encoder $g(y)$.
The model optimizes a contrastive cross-entropy loss that encourages a high dot-product $f(x) \cdot g(y)$ if the image $x$ is paired with the given caption $y$, or a low dot-product if the image and caption correspond to different parts in the training data.
The denoising diffusion process can be perturbed with the gradient of the dot product of the image and caption encodings with respect to the image as follows:
\begin{equation}
    \hat{\epsilon}_{\theta}' (x_t | c) = \epsilon_{\theta} (x_t|c) - \sqrt{1 - \bar{\alpha}_{t}} \nabla_{x_t} (f(x_t) \cdot g(y))
\end{equation}
As they feed noised samples $x_t$ to the CLIP model, so they must train CLIP on noised images $x_t$ to obtain correct gradient in the reverse process.
Their experiments show that classifier-free guidance yields higher quality images than CLIP guidance.

In this work, we show that CLIP guidance can be combined with the classifier-free guidance to produce much higher quality samples. 
Moreover, any pre-trained image-text matching models can be effectively utilized without training on noised images by several crucial improvements.

\begin{figure}
    \centering
    \includegraphics[width=5.5in]{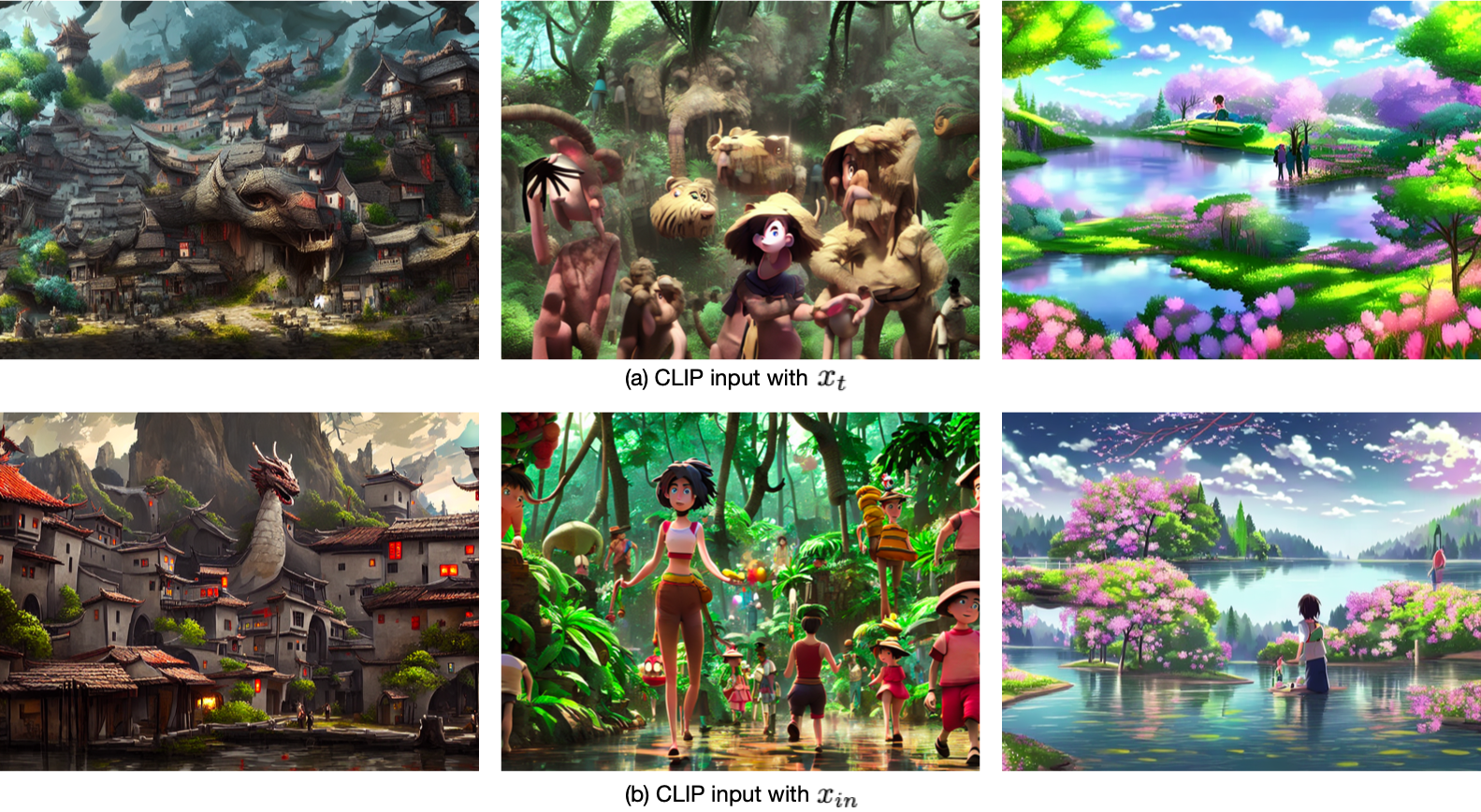}
    \caption{Comparing samples between CLIP inputs with $x_t$ and the modified $x_{in}$. The results show that feeding CLIP guidance with modified $x_{in}$ is critical to improve the sample quality.}
    \label{fig:clip_input}
\end{figure}

\paragraph{Modifying the CLIP inputs.}
Noised samples $x_t$ are more like Gaussian noises that capture very few meaningful semantics in earlier sampling steps, especially when $t$ is near $T$.
It's difficult to obtain effective guidance by directly feeding $x_t$ into the image-text matching model which are pre-trained on natural images, such as CLIP.
Although further training a noisy-aware CLIP help alleviate this problem, it still has two big drawbacks: (1) training a noisy-aware CLIP on noised images $x_t$ is time and computation costly. Many existing pre-trained image-text matching models cannot be utilized directly; (2) noised images $x_t$ when $t$ is near $T$ is close to pure Gaussian noise, which is difficult to be aligned to corresponding text to give meaningful guidance to diffusion models.

To get more meaningful inputs to CLIP models, we instead first predict the final images $\hat{x}_0 = \frac{1}{\sqrt{\bar{\alpha}_t}} (x_t - \sqrt{1 - \bar{\alpha}_t} \epsilon_t)$  (\ref{eq_x0}), and extrapolate the noised images $x_t$ with the prediction $\hat{x}_0$, which can be formulated as follows:
\begin{equation}
    x_{in} = \sqrt{1 - \bar{\alpha}_t} \hat{x}_0 + (1 - \sqrt{1 - \bar{\alpha}_t}) x_t
\end{equation}
\noindent where $\sqrt{1 - \bar{\alpha}_t} \in [1, 0]$ when $t \in [T, 1]$. 
Thus, in earlier steps that $t$ is close to $T$, the CLIP inputs $x_{in}$ is dominated by $\hat{x}_0$ and gradually close to $x_t$.
Our experiments show that this modification is critical for the effectiveness of CLIP guidance. Figure~\ref{fig:clip_input} shows some comparing examples.

\begin{figure}[htbp]
\begin{minipage}[l]{0.38\textwidth}
\vspace{0pt}
\includegraphics[width=2.3in]{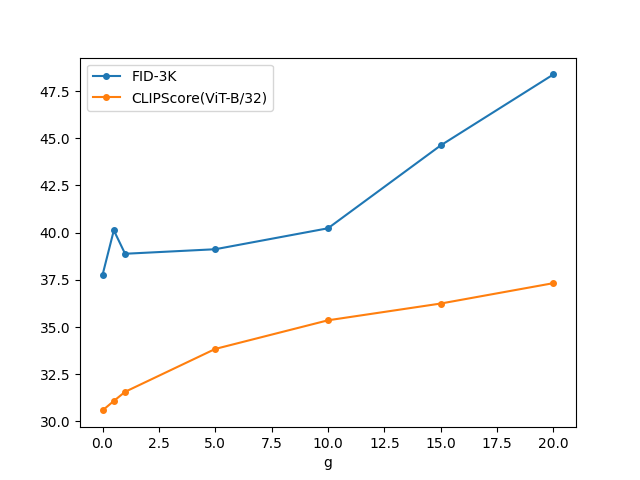}
\caption{Comparing the performance of different CLIP guidance weight $g$. We set $g=10.0$ in our experiments to obtain a good balance between image-text alignment and image fidelity.}
\label{fig:cond_weight}
\end{minipage}
\begin{minipage}[r]{0.58\textwidth}
\vspace{0pt}
\includegraphics[width=3.3in]{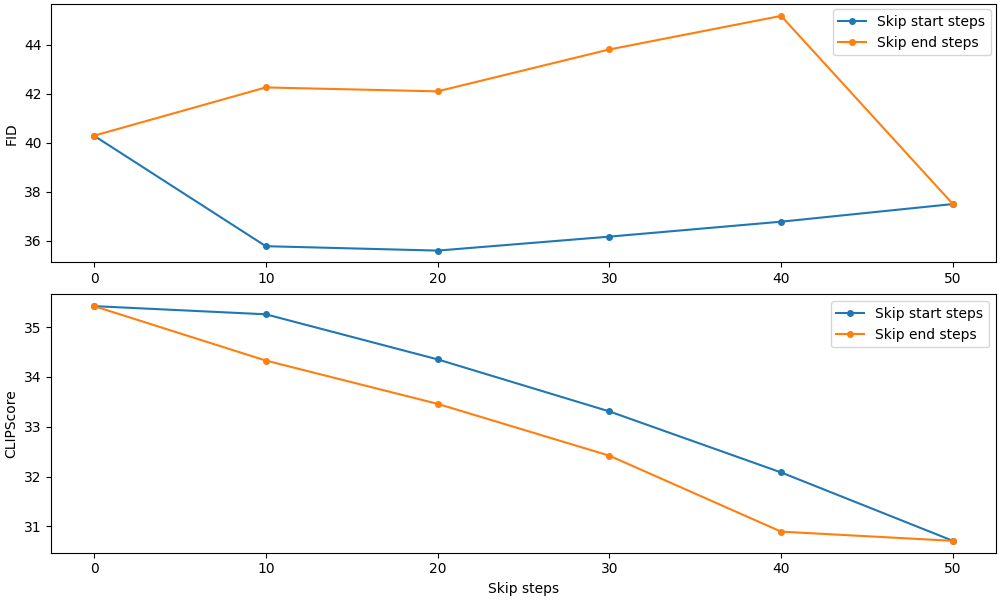}
\caption{Comparing the performance of different skip scheduling methods. We adopt DDIM sampling approach with totally 50 diffusion steps by setting the noise to $0$.}
\label{fig:skip_scheduling}
\end{minipage}
\end{figure}

\paragraph{Combining with classifier-free guidance.}
The classifier-free guidance is very effective to improve image-text alignment for text-conditional diffusion models.
However, increasing the classifier-free guidance weight will also damages image fidelity producing highly saturated and unnatural images.
Image-text matching models that pre-trained on large-volumes of paired image-text data capture rich cross-modal semantics, style and aesthetics.
Combining Image-text matching guidance and classifier-free guidance can simultaneously help improve sample fidelity, aesthetics and image-text alignment.
Thus, the denoising diffusion process can be formulated as follows:
\begin{equation}
    \hat{\epsilon}_{\theta}'' (x_t | c) = \epsilon_{\theta}(x_t|\emptyset) + s \cdot (\epsilon_{\theta} (x_t|c) - \epsilon_{\theta}(x_t|\emptyset)) - g \sqrt{1 - \bar{\alpha}_{t}} \nabla_{x_t} (f(x_{in}) \cdot g(y))
\label{eq_combine}
\end{equation}
\noindent where $g \ge 0$ is the image-text matching guidance weight. 
Figure~\ref{fig:cond_weight} compares the performance of different $g$. In our experiments, we set $g=10.0$ to achieve better performance on text-to-image generation.

\paragraph{Diverse guidance schedules.}
As samples $x_t$ at earlier diffusion steps capture very few meaningful semantics, the image-text matching guidance weight should be weaker or even skipped at earlier steps, and be increased as $T$ close to $1$. We compare several different settings that skip different steps, and show that skipping a few earlier diffusion steps can effectively improve the sample quality.
Figure~\ref{fig:skip_scheduling} compares the performance of different schedule settings.
The results show that skipping a few earlier steps will increase the sample fidelity, however, skipping too many steps will decrease both sample fidelity and alignment. Thus, in the following experiments, we skip 10 steps of CLIP guidance in the start with totally 50 DDIM steps.

\begin{figure}
    \centering
    \includegraphics[width=5in]{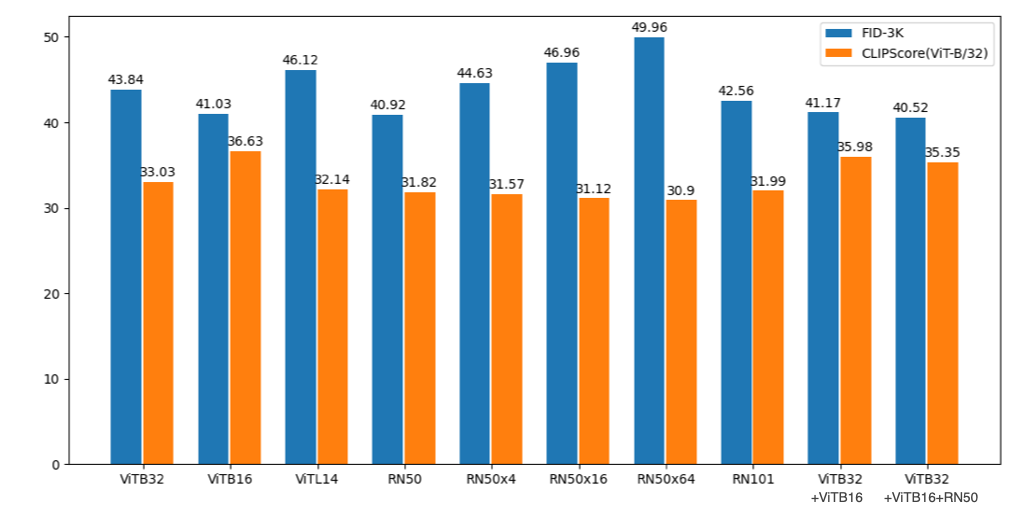}
    \caption{Comparing the performance of utilizing different CLIP models for cross-modal guidance.}
    \label{fig:comparing_clips}
\end{figure}

\paragraph{Integrate multiple image-text matching models.}
Image-text matching has achieved great improvements based on image-text contrastive learning.
Many vision-language pre-training models have been pre-trained on different datasets with different settings, which may capture different aspects of cross-modal semantics.
It's very intuitive to integrate multiple image-text matching models to capture richer cross-modal semantics.
We denote $F$ as a set of image-text matching models.
Then Equation~\ref{eq_combine} can be modified as follows:
\begin{equation}
    \hat{\epsilon}_{\theta}'' (x_t | c) = \epsilon_{\theta}(x_t|\emptyset) + s \cdot (\epsilon_{\theta} (x_t|c) - \epsilon_{\theta}(x_t|\emptyset)) - \frac{g \sqrt{1 - \bar{\alpha}_{t}}}{|F|} \sum_{f \in F} \nabla_{x_t} (f(x_{in}) \cdot g(y))
\label{eq_multiclip}
\end{equation}
Figure~\ref{fig:comparing_clips} shows the results of using different image-text matching models in our model. The results show that different image-text matching models have different performance on sample fidelity (i.e. FID) and caption similarity (i.e. CLIPScore), composing them together can achieve better performance by giving more comprehensive cross-modal guidance.

\subsection{Training and Inference}
We only need to train the text-conditional diffusion model, and directly incorporate pre-trained image-text matching models during inference.
We adopt the U-Net architecture from~\citet{ho2020denoising} for our text-to-image diffusion model.
The network is conditioned on text embeddings from the text encoder by two aspects: a pooled embedding vector added to the diffusion timestep embedding, and cross attention over the entire sequence of text embeddings.
We train on a combination of internal Chinese datasets, with $\approx 600M$ image-text pairs, and the publicly available Laion dataset, with $\approx 400M$ image-text pairs.
The Laion dataset has been translated into Chinese.

During inference, existing pre-trained image-text matching models can all be incorporated into the diffusion process, such as CLIP~\citep{radford2021learning}, UNIMO~\citep{li2020unimo}, and Ernie-vil~\citep{yu2021ernie}.
In our experiments, we utilize CLIP to evaluate the performance of our models, as it is most widely used in the community.
As shown in Figure~\ref{fig:comparing_clips}, different versions of CLIP models have different performance, we compose the ViTB32, ViTB16 and RN50 models to achieve a good trade-off between performance and efficiency.
We also combine the classifier-free guidance and image-text matching guidance by setting the classifier-free guidance scale $s=8.0$ and the image-text matching guidance weight $g=10.0$.
We adopt DDIM~\citep{song2020denoising} sampling approach with totally 50 diffusion steps by setting the noise to $0$.
During sampling, we skip the first 10 steps without using the image-text matching guidance and apply it in the following 40 diffusion steps, as discussed in Subsection~\ref{subs:img2txt_matching}.

\begin{figure}
    \centering
    \includegraphics[width=5.5in]{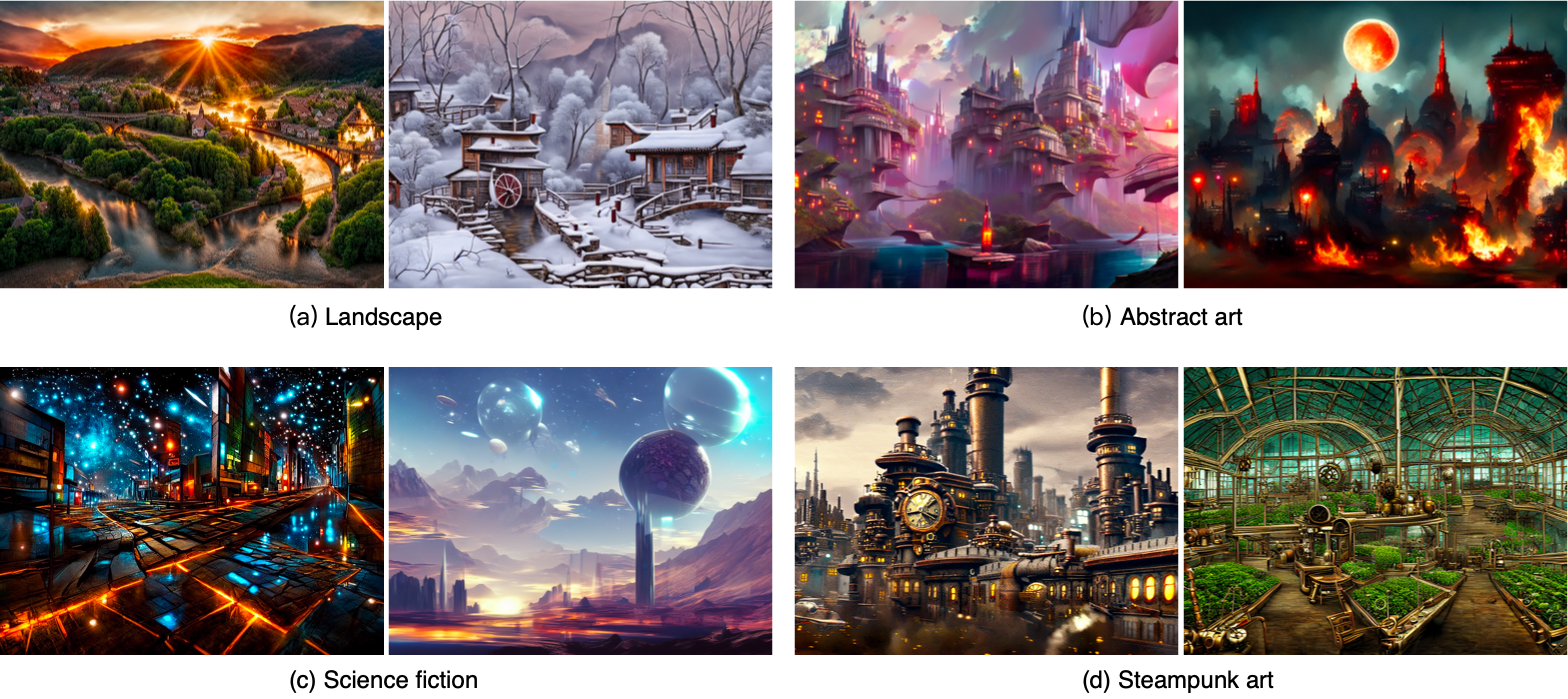}
    \caption{UPainting samples for different categories of complex-scene prompts from UniBench.}
    \label{fig:complex_scene_examples}
\end{figure}

\begin{figure}
    \centering
    \includegraphics[width=5.5in]{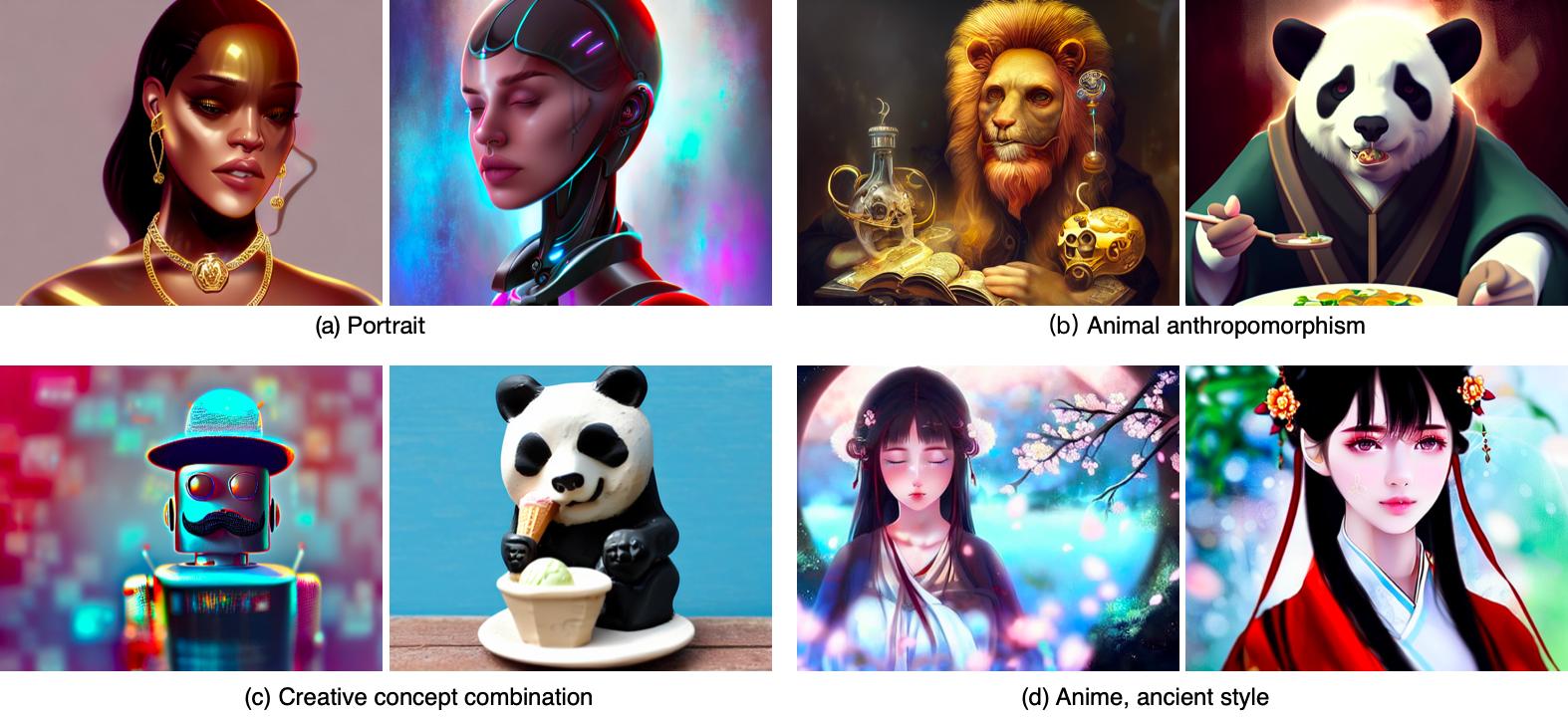}
    \caption{UPainting samples for different categories of simple scene prompts from UniBench.}
    \label{fig:small_scene_examples}
\end{figure}

\section{Evaluation Benchmark}
\label{benchmark}
COCO~\citep{lin2014microsoft} is the standard benchmark for evaluating text-to-image generation models, which mainly composition queries for simple scene images.
The main used automatic metrics are FID~\citep{heusel2017gans} to measure image fidelity and CLIPscore~\citep{hessel2021clipscore} to measure image-text alignment.
We compare UPainting with existing models in both the supervised and the zero-shot settings.
Consistent with previous works, we randomly sample 30K prompts from the COCO validation set as evaluation data.
For model analysis in Subsection~\ref{subs:img2txt_matching}, we randomly sample 3K prompts from the COCO for efficiency and report both FID and CLIPscore. Besides COCO, several other benchmarks have also been proposed to systematically evaluate the performance of different text-to-image models, such as PaintSkills~\citep{cho2022dall}, DrawBench~\citep{saharia2022photorealistic} and PartiPrompts~\citep{yu2022scaling}.
These benchmarks don't have reference images, so it's difficult to make automatic evaluation on them.
And, the prompts of these benchmarks are also mainly for simple-scene hyper-realistic images.

\paragraph{UniBench} 
To more comprehensively compare capabilities of different text-to-image generation models on generating both simple-scene and complex-scene images, we further propose a more general evaluation benchmark, UniBench.
UniBench contains both prompts for simple-scene images and complex-scene images as shown in Figure~\ref{fig:leading_samples} and Figure~\ref{fig:more_samples}, all selected from DALL-E 2~\citep{ramesh2022hierarchical}, Stable Diffusion~\footnote{https://lexica.art/}, Reddit~\footnote{https://www.reddit.com/} and online queries from YiGe\footnote{https://yige.baidu.com/}.
UniBench contains a total of 200 prompts, including 100 simple-scene queries and 100 complex-scene queries. It provides both Chinese and English languages for fair comparison of Chinese and English models.
UniBench can be used to measure model capabilities across various categories and challenge aspects, some of which are shown in Figure~\ref{fig:complex_scene_examples} and Figure~\ref{fig:small_scene_examples}.

\section{Experiments}
\label{experiments}
We evaluate UPainting in both automatic and human evaluation settings. For all our experiments, the images are random generated samples from UPainting without post-processing or re-ranking.

\begin{table}[t!]
\centering
\caption{Evaluation results on MS-COCO 256x256 FID-30K.}
\begin{tabular}{l c}
\hline
Model & FID Score (30K) \\
\hline
AttnGAN\citep{xu2018attngan} & 35.49  \\
DM-GAN\citep{zhu2019dm} & 32.64  \\
DF-GAN\citep{tao2020df} & 21.42 \\
DM-GAN + CL\citep{ye2021improving} & 20.79 \\
XMC-GAN\citep{zhang2021cross} & 9.33 \\
LAFITE\citep{zhou2021lafite} & 8.12 \\
Mask-A-Scene\citep{gafni2022make} & 7.55 \\
\hline
DALL-E\citep{ramesh2021zero} & 17.89   \\
GLIDE\citep{nichol2021glide} & 12.24  \\
DALL-E 2\citep{ramesh2022hierarchical} & 10.39  \\
Imagen\citep{saharia2022photorealistic} & \textbf{7.27} \\
Stable Diffusion\citep{rombach2022high} & 14.24 \\
\textbf{UPainting (Our work)} & 8.34  \\
\hline
\end{tabular}
\label{COCO}
\end{table}

\subsection{Automatic Evaluation}
We use FID score to evaluate UPainting on the COCO validation set, similar to~\citet{saharia2022photorealistic,ramesh2022hierarchical}.
Consistent with previous works, we report FID-30K, for which 30K prompts are drawn randomly from the validation set, and the model samples generated on these prompts are compared with reference images from the full validation set.
The captions are translated into Chinese.
Although there are mistakes when translating English captions into Chinese, UPainting outperforms other recent works such as DALL-E 2 and Stable Diffusion, as shown in Table~\ref{COCO}.
Note that FID is not a perfect metric to evaluate UPainting because the advantage of UPainting is to generate both simple scene and complex scene images for complex prompts (including sophisticated art styles, aesthetic features, etc.), but the prompts of COCO are mainly specific descriptions of simple scene images.

\subsection{Human Evaluation}
UniBench doesn't have golden reference images, so we apply human evaluation on this benchmark.
Two human raters are asked to compare UPainting with two strong baselines side by side independently: (1) Disco Diffusion~\citep{Crowson2021dd512}, good at generating imagism images of complex scenes; and (2) Stable Diffusion~\citep{rombach2022high}, one of the most powerful text-to-image diffusion model.
For each prompt, the raters choose to prefer Model A, Model B, or are indifferent for both image fidelity and image-text alignment. 

\begin{figure}[tp]
    \centering
    \includegraphics[width=5.5in]{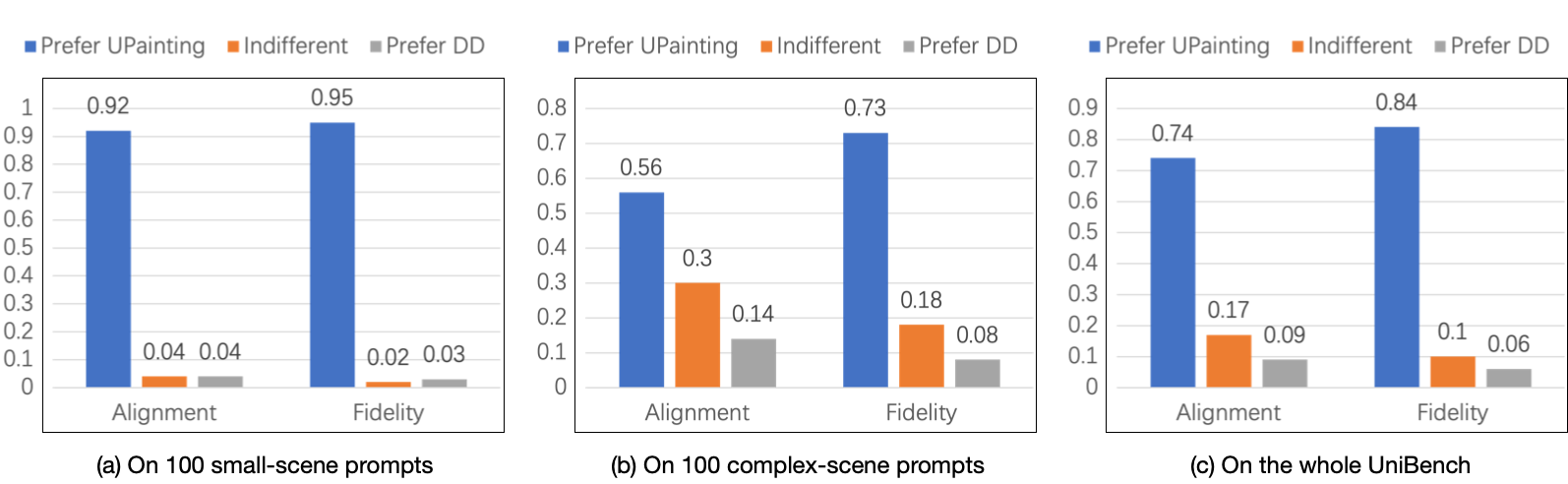}
    \caption{The comparison of user preference rates for image-text alignment and image fidelity between UPainting and Disco Diffusion (i.e. DD) on UniBench.}
    \label{fig:compare_dd}
\end{figure}

\paragraph{Comparison to Disco Diffusion.} Disco Diffusion (i.e. DD) is one of the most popular open released cross-modal guided image generation model. DD applies CLIP models to guide an unconditional diffusion model for text-conditional image generation.
Similar to our model, DD guides the diffusion process iteratively through CLIP gradients.
It's good at complex-scene imagism image generation, but difficult to generate details like specific objects, actions, and attributes.
Figure~\ref{fig:compare_dd} shows the comparison results between UPainting and Disco Diffusion, which demonstrate that human raters exceedingly prefer UPainting in both image-text alignment and image fidelity.
Especially in the generation of simple-scene images that require generating concrete objects, UPainting is better than DD in more than 90\% of the prompts.
On complex-scene prompts, UPainting can also generate images with better aesthetics and accurate details.
Some comparison examples are listed in Figure~\ref{fig:comparison_small_scene} and Figure~\ref{fig:comparison_complex_scene}.

\begin{figure}
    \centering
    \includegraphics[width=5.5in]{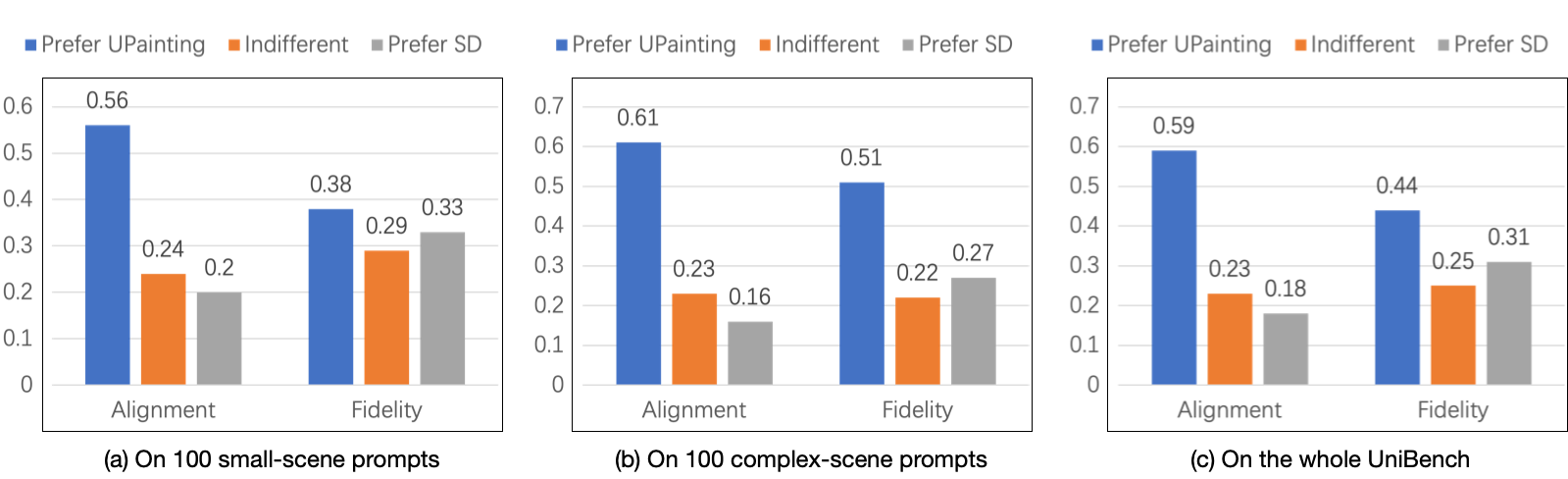}
    \caption{The comparison of user preference rates for image-text alignment and image fidelity between UPainting and Stable Diffusion (i.e. SD) on UniBench.}
    \label{fig:compare_sd}
\end{figure}

\paragraph{Comparison to Stable Diffusion.}
Stable Diffusion is one of the state-of-the-art text conditional diffusion model.
Figure~\ref{fig:compare_sd} shows the comparison results between UPainting and Stable Diffusion (i.e. SD), which show that human raters much prefer UPainting over SD in both image-text alignment and image fidelity.
The advantages of UPainting over SD are much larger on the complex scene prompts.
The images generated by UPainting generally have better caption similarity and aesthetics.
Some examples are also shown in Figure~\ref{fig:comparison_small_scene} and Figure~\ref{fig:comparison_complex_scene}.
The results demonstrate the effectiveness of combining the power of large pre-trained transformer language model in language understanding and pre-trained image-text matching models in capturing cross-modal semantics and styles.


\section{Related Work}
\label{related_work}
Text-conditional image generation has achieved great progress during recent years. 
The main technique routes of text-to-image generation include three stages: GAN-based~\citep{xu2018attngan,zhu2019dm,tao2020df,zhang2021cross}, auto-regressive Transformer-based~\citep{gafni2022make,ramesh2021zero,zhang2021ernie,huang2022vlg,ding2021cogview}, and diffusion-based~\citep{ho2020denoising,dhariwal2021diffusion,nichol2021glide,ramesh2022hierarchical,saharia2022photorealistic,rombach2022high}.
Many earlier works have trained GANs on publicly available image captioning datasets to produce text-conditional images.
These works mainly focus on specific domains of image generation.
Motivated by the success of GPT-3~\citep{brown2020language}, auto-regressive transformers later have been extensively trained on sequences of text tokens followed by image tokens by adapting the VQ-VAE~\citep{van2017neural} approach.
The most well-known approach, DALL-E~\citep{ramesh2021zero}, attracts great attention of the community.
Most recently, diffusion model brings wide success to image synthesis.
\citet{nichol2021glide} proposes the first text-conditional image-to-text diffusion model GLIDE.
Based on GLIDE, Imagen~\citep{saharia2022photorealistic} validates the effectiveness of utilizing a large pre-trained Transformer language model as text encoder, which achieves state-of-the art FID on the COCO dataset.
Instead of applying diffusion on the pixel space, \citet{rombach2022high} propose a latent diffusion model that applies diffusion in the latent space of powerful pre-trained auto-encoders, which effectively reduce the computational resources of diffusion training.
These models mainly focus on hyper-realistic generation of simple scenes, and usually show low levels of detail for some complex scenes~\citep{ramesh2022hierarchical}.

Image-text matching models, typically CLIPs, have been extensively utilized to steer image generation models towards text conditions~\citep{liu2021more}.
Based on CLIP, \citet{ramesh2022hierarchical} propose a two-stage diffusion model DALL-E 2, where a prior generates a CLIP image embedding given a text caption, and a decoder generates an image conditioned on the image embedding.
Through incorporating CLIP latents into diffusion image generation, DALL-E 2 greatly improves the diversity of generated samples.
\citet{nichol2021glide} propose to train a noise-aware CLIP model on noised images and guide a text-conditional diffusion model by its gradients.
However, their experiments show classifier-free guidance works more favorably than CLIP guidance for text conditional image generation.
Moreover, training a noise-aware CLIP model is time and resource costly.
Existing pre-trained CLIP models cannot be directly utilized.
\citet{Crowson2021dd256,Crowson2021dd512} use an unnoised CLIP model to guide unconditional or class-conditional diffusion models.
However, it mainly targets abstract scene images without concrete details.
In this paper, we propose an effective model that combines the pre-trained Transformer language model and pre-trained image-text matching models, which greatly improves both the caption similarity and image fidelity for both simple scene and complex scene image generation.


\section{Conclusion}
In this paper, we systematically study the problem of text-conditional image generation for both simple and complex scenes, and propose a simple yet effective method to unify them.
We find that effectively combining cross-modal matching models with pre-trained transformer language models can greatly improve sample fidelity and image-text alignment for diffusion image generation, which gives the model a general ability to generate images for both simple and complex scenes.
To more comprehensively compare different text-to-image generation models, we also propose a comprehensive and challenging evaluation benchmark for both simple and complex scene image generation.
On this benchmark, UPainting greatly outperforms other strong models such as Stable Diffusion and Disco Diffusion, on both image-text alignment and image fidelity.

\bibliographystyle{plainnat}
\bibliography{UPainting}

\begin{thebibliography}{37}
\providecommand{\natexlab}[1]{#1}
\providecommand{\url}[1]{\texttt{#1}}
\expandafter\ifx\csname urlstyle\endcsname\relax
  \providecommand{\doi}[1]{doi: #1}\else
  \providecommand{\doi}{doi: \begingroup \urlstyle{rm}\Url}\fi

\bibitem[Brown et~al.(2020)Brown, Mann, Ryder, Subbiah, Kaplan, Dhariwal,
  Neelakantan, Shyam, Sastry, Askell, et~al.]{brown2020language}
Tom Brown, Benjamin Mann, Nick Ryder, Melanie Subbiah, Jared~D Kaplan, Prafulla
  Dhariwal, Arvind Neelakantan, Pranav Shyam, Girish Sastry, Amanda Askell,
  et~al.
\newblock Language models are few-shot learners.
\newblock \emph{Advances in neural information processing systems},
  33:\penalty0 1877--1901, 2020.

\bibitem[Cho et~al.(2022)Cho, Zala, and Bansal]{cho2022dall}
Jaemin Cho, Abhay Zala, and Mohit Bansal.
\newblock Dall-eval: Probing the reasoning skills and social biases of
  text-to-image generative transformers.
\newblock \emph{arXiv preprint arXiv:2202.04053}, 2022.

\bibitem[Crowson(2021{\natexlab{a}})]{Crowson2021dd256}
Katherine Crowson.
\newblock Clip guided diffusion hq 256x256.
\newblock
  \emph{https://colab.research.google.com/drive/12a\_Wrfi2\_gwwAuN3VvMTwVMz9TfqctNj},
  2021{\natexlab{a}}.

\bibitem[Crowson(2021{\natexlab{b}})]{Crowson2021dd512}
Katherine Crowson.
\newblock Clip guided diffusion 512x512, secondary model method.
\newblock
  \emph{https://twitter.com/RiversHaveWings/status/1462859669454536711},
  2021{\natexlab{b}}.

\bibitem[Devlin et~al.(2018)Devlin, Chang, Lee, and Toutanova]{devlin2018bert}
Jacob Devlin, Ming-Wei Chang, Kenton Lee, and Kristina Toutanova.
\newblock Bert: Pre-training of deep bidirectional transformers for language
  understanding.
\newblock \emph{arXiv preprint arXiv:1810.04805}, 2018.

\bibitem[Dhariwal and Nichol(2021)]{dhariwal2021diffusion}
Prafulla Dhariwal and Alexander Nichol.
\newblock Diffusion models beat gans on image synthesis.
\newblock \emph{Advances in Neural Information Processing Systems},
  34:\penalty0 8780--8794, 2021.

\bibitem[Ding et~al.(2021)Ding, Yang, Hong, Zheng, Zhou, Yin, Lin, Zou, Shao,
  Yang, et~al.]{ding2021cogview}
Ming Ding, Zhuoyi Yang, Wenyi Hong, Wendi Zheng, Chang Zhou, Da~Yin, Junyang
  Lin, Xu~Zou, Zhou Shao, Hongxia Yang, et~al.
\newblock Cogview: Mastering text-to-image generation via transformers.
\newblock \emph{Advances in Neural Information Processing Systems},
  34:\penalty0 19822--19835, 2021.

\bibitem[Gafni et~al.(2022)Gafni, Polyak, Ashual, Sheynin, Parikh, and
  Taigman]{gafni2022make}
Oran Gafni, Adam Polyak, Oron Ashual, Shelly Sheynin, Devi Parikh, and Yaniv
  Taigman.
\newblock Make-a-scene: Scene-based text-to-image generation with human priors.
\newblock \emph{arXiv preprint arXiv:2203.13131}, 2022.

\bibitem[Hessel et~al.(2021)Hessel, Holtzman, Forbes, Bras, and
  Choi]{hessel2021clipscore}
Jack Hessel, Ari Holtzman, Maxwell Forbes, Ronan~Le Bras, and Yejin Choi.
\newblock Clipscore: A reference-free evaluation metric for image captioning.
\newblock \emph{arXiv preprint arXiv:2104.08718}, 2021.

\bibitem[Heusel et~al.(2017)Heusel, Ramsauer, Unterthiner, Nessler, and
  Hochreiter]{heusel2017gans}
Martin Heusel, Hubert Ramsauer, Thomas Unterthiner, Bernhard Nessler, and Sepp
  Hochreiter.
\newblock Gans trained by a two time-scale update rule converge to a local nash
  equilibrium.
\newblock \emph{Advances in neural information processing systems}, 30, 2017.

\bibitem[Ho and Salimans(2022)]{ho2022classifier}
Jonathan Ho and Tim Salimans.
\newblock Classifier-free diffusion guidance.
\newblock \emph{arXiv preprint arXiv:2207.12598}, 2022.

\bibitem[Ho et~al.(2020)Ho, Jain, and Abbeel]{ho2020denoising}
Jonathan Ho, Ajay Jain, and Pieter Abbeel.
\newblock Denoising diffusion probabilistic models.
\newblock \emph{Advances in Neural Information Processing Systems},
  33:\penalty0 6840--6851, 2020.

\bibitem[Huang et~al.(2022)Huang, Niu, Liu, Xiao, and Wu]{huang2022vlg}
Luyang Huang, Guocheng Niu, Jiachen Liu, Xinyan Xiao, and Hua Wu.
\newblock Du-vlg: Unifying vision-and-language generation via dual
  sequence-to-sequence pre-training.
\newblock \emph{arXiv preprint arXiv:2203.09052}, 2022.

\bibitem[Li et~al.(2021)Li, Selvaraju, Gotmare, Joty, Xiong, and
  Hoi]{li2021align}
Junnan Li, Ramprasaath Selvaraju, Akhilesh Gotmare, Shafiq Joty, Caiming Xiong,
  and Steven Chu~Hong Hoi.
\newblock Align before fuse: Vision and language representation learning with
  momentum distillation.
\newblock \emph{Advances in neural information processing systems},
  34:\penalty0 9694--9705, 2021.

\bibitem[Li et~al.(2020)Li, Gao, Niu, Xiao, Liu, Liu, Wu, and
  Wang]{li2020unimo}
Wei Li, Can Gao, Guocheng Niu, Xinyan Xiao, Hao Liu, Jiachen Liu, Hua Wu, and
  Haifeng Wang.
\newblock Unimo: Towards unified-modal understanding and generation via
  cross-modal contrastive learning.
\newblock \emph{arXiv preprint arXiv:2012.15409}, 2020.

\bibitem[Li et~al.(2022)Li, Gao, Niu, Xiao, Liu, Liu, Wu, and
  Wang]{li2022unimo}
Wei Li, Can Gao, Guocheng Niu, Xinyan Xiao, Hao Liu, Jiachen Liu, Hua Wu, and
  Haifeng Wang.
\newblock Unimo-2: End-to-end unified vision-language grounded learning.
\newblock \emph{arXiv preprint arXiv:2203.09067}, 2022.

\bibitem[Lin et~al.(2014)Lin, Maire, Belongie, Hays, Perona, Ramanan,
  Doll{\'a}r, and Zitnick]{lin2014microsoft}
Tsung-Yi Lin, Michael Maire, Serge Belongie, James Hays, Pietro Perona, Deva
  Ramanan, Piotr Doll{\'a}r, and C~Lawrence Zitnick.
\newblock Microsoft coco: Common objects in context.
\newblock In \emph{European conference on computer vision}, pages 740--755.
  Springer, 2014.

\bibitem[Liu et~al.(2021)Liu, Park, Azadi, Zhang, Chopikyan, Hu, Shi, Rohrbach,
  and Darrell]{liu2021more}
Xihui Liu, Dong~Huk Park, Samaneh Azadi, Gong Zhang, Arman Chopikyan, Yuxiao
  Hu, Humphrey Shi, Anna Rohrbach, and Trevor Darrell.
\newblock More control for free! image synthesis with semantic diffusion
  guidance.
\newblock \emph{arXiv preprint arXiv:2112.05744}, 2021.

\bibitem[Nichol et~al.(2021)Nichol, Dhariwal, Ramesh, Shyam, Mishkin, McGrew,
  Sutskever, and Chen]{nichol2021glide}
Alex Nichol, Prafulla Dhariwal, Aditya Ramesh, Pranav Shyam, Pamela Mishkin,
  Bob McGrew, Ilya Sutskever, and Mark Chen.
\newblock Glide: Towards photorealistic image generation and editing with
  text-guided diffusion models.
\newblock \emph{arXiv preprint arXiv:2112.10741}, 2021.

\bibitem[Radford et~al.(2021)Radford, Kim, Hallacy, Ramesh, Goh, Agarwal,
  Sastry, Askell, Mishkin, Clark, et~al.]{radford2021learning}
Alec Radford, Jong~Wook Kim, Chris Hallacy, Aditya Ramesh, Gabriel Goh,
  Sandhini Agarwal, Girish Sastry, Amanda Askell, Pamela Mishkin, Jack Clark,
  et~al.
\newblock Learning transferable visual models from natural language
  supervision.
\newblock In \emph{International Conference on Machine Learning}, pages
  8748--8763. PMLR, 2021.

\bibitem[Raffel et~al.(2020)Raffel, Shazeer, Roberts, Lee, Narang, Matena,
  Zhou, Li, Liu, et~al.]{raffel2020exploring}
Colin Raffel, Noam Shazeer, Adam Roberts, Katherine Lee, Sharan Narang, Michael
  Matena, Yanqi Zhou, Wei Li, Peter~J Liu, et~al.
\newblock Exploring the limits of transfer learning with a unified text-to-text
  transformer.
\newblock \emph{J. Mach. Learn. Res.}, 21\penalty0 (140):\penalty0 1--67, 2020.

\bibitem[Ramesh et~al.(2021)Ramesh, Pavlov, Goh, Gray, Voss, Radford, Chen, and
  Sutskever]{ramesh2021zero}
Aditya Ramesh, Mikhail Pavlov, Gabriel Goh, Scott Gray, Chelsea Voss, Alec
  Radford, Mark Chen, and Ilya Sutskever.
\newblock Zero-shot text-to-image generation.
\newblock In \emph{International Conference on Machine Learning}, pages
  8821--8831. PMLR, 2021.

\bibitem[Ramesh et~al.(2022)Ramesh, Dhariwal, Nichol, Chu, and
  Chen]{ramesh2022hierarchical}
Aditya Ramesh, Prafulla Dhariwal, Alex Nichol, Casey Chu, and Mark Chen.
\newblock Hierarchical text-conditional image generation with clip latents.
\newblock \emph{arXiv preprint arXiv:2204.06125}, 2022.

\bibitem[Rombach et~al.(2022)Rombach, Blattmann, Lorenz, Esser, and
  Ommer]{rombach2022high}
Robin Rombach, Andreas Blattmann, Dominik Lorenz, Patrick Esser, and Bj{\"o}rn
  Ommer.
\newblock High-resolution image synthesis with latent diffusion models.
\newblock In \emph{Proceedings of the IEEE/CVF Conference on Computer Vision
  and Pattern Recognition}, pages 10684--10695, 2022.

\bibitem[Saharia et~al.(2022)Saharia, Chan, Saxena, Li, Whang, Denton,
  Ghasemipour, Ayan, Mahdavi, Lopes, et~al.]{saharia2022photorealistic}
Chitwan Saharia, William Chan, Saurabh Saxena, Lala Li, Jay Whang, Emily
  Denton, Seyed Kamyar~Seyed Ghasemipour, Burcu~Karagol Ayan, S~Sara Mahdavi,
  Rapha~Gontijo Lopes, et~al.
\newblock Photorealistic text-to-image diffusion models with deep language
  understanding.
\newblock \emph{arXiv preprint arXiv:2205.11487}, 2022.

\bibitem[Song et~al.(2020)Song, Meng, and Ermon]{song2020denoising}
Jiaming Song, Chenlin Meng, and Stefano Ermon.
\newblock Denoising diffusion implicit models.
\newblock \emph{arXiv preprint arXiv:2010.02502}, 2020.

\bibitem[Sun et~al.(2020)Sun, Wang, Li, Feng, Tian, Wu, and Wang]{sun2020ernie}
Yu~Sun, Shuohuan Wang, Yukun Li, Shikun Feng, Hao Tian, Hua Wu, and Haifeng
  Wang.
\newblock Ernie 2.0: A continual pre-training framework for language
  understanding.
\newblock In \emph{Proceedings of the AAAI Conference on Artificial
  Intelligence}, volume~34, pages 8968--8975, 2020.

\bibitem[Tao et~al.(2020)Tao, Tang, Wu, Sebe, Jing, Wu, and Bao]{tao2020df}
Ming Tao, Hao Tang, Songsong Wu, Nicu Sebe, Xiao-Yuan Jing, Fei Wu, and Bingkun
  Bao.
\newblock Df-gan: Deep fusion generative adversarial networks for text-to-image
  synthesis.
\newblock \emph{arXiv preprint arXiv:2008.05865}, 2020.

\bibitem[Van Den~Oord et~al.(2017)Van Den~Oord, Vinyals, et~al.]{van2017neural}
Aaron Van Den~Oord, Oriol Vinyals, et~al.
\newblock Neural discrete representation learning.
\newblock \emph{Advances in neural information processing systems}, 30, 2017.

\bibitem[Xu et~al.(2018)Xu, Zhang, Huang, Zhang, Gan, Huang, and
  He]{xu2018attngan}
Tao Xu, Pengchuan Zhang, Qiuyuan Huang, Han Zhang, Zhe Gan, Xiaolei Huang, and
  Xiaodong He.
\newblock Attngan: Fine-grained text to image generation with attentional
  generative adversarial networks.
\newblock In \emph{Proceedings of the IEEE conference on computer vision and
  pattern recognition}, pages 1316--1324, 2018.

\bibitem[Ye et~al.(2021)Ye, Yang, Takac, Sunderraman, and Ji]{ye2021improving}
Hui Ye, Xiulong Yang, Martin Takac, Rajshekhar Sunderraman, and Shihao Ji.
\newblock Improving text-to-image synthesis using contrastive learning.
\newblock \emph{arXiv preprint arXiv:2107.02423}, 2021.

\bibitem[Yu et~al.(2021)Yu, Tang, Yin, Sun, Tian, Wu, and Wang]{yu2021ernie}
Fei Yu, Jiji Tang, Weichong Yin, Yu~Sun, Hao Tian, Hua Wu, and Haifeng Wang.
\newblock Ernie-vil: Knowledge enhanced vision-language representations through
  scene graphs.
\newblock In \emph{Proceedings of the AAAI Conference on Artificial
  Intelligence}, volume~35, pages 3208--3216, 2021.

\bibitem[Yu et~al.(2022)Yu, Xu, Koh, Luong, Baid, Wang, Vasudevan, Ku, Yang,
  Ayan, et~al.]{yu2022scaling}
Jiahui Yu, Yuanzhong Xu, Jing~Yu Koh, Thang Luong, Gunjan Baid, Zirui Wang,
  Vijay Vasudevan, Alexander Ku, Yinfei Yang, Burcu~Karagol Ayan, et~al.
\newblock Scaling autoregressive models for content-rich text-to-image
  generation.
\newblock \emph{arXiv preprint arXiv:2206.10789}, 2022.

\bibitem[Zhang et~al.(2021{\natexlab{a}})Zhang, Koh, Baldridge, Lee, and
  Yang]{zhang2021cross}
Han Zhang, Jing~Yu Koh, Jason Baldridge, Honglak Lee, and Yinfei Yang.
\newblock Cross-modal contrastive learning for text-to-image generation.
\newblock In \emph{Proceedings of the IEEE/CVF conference on computer vision
  and pattern recognition}, pages 833--842, 2021{\natexlab{a}}.

\bibitem[Zhang et~al.(2021{\natexlab{b}})Zhang, Yin, Fang, Li, Duan, Wu, Sun,
  Tian, Wu, and Wang]{zhang2021ernie}
Han Zhang, Weichong Yin, Yewei Fang, Lanxin Li, Boqiang Duan, Zhihua Wu,
  Yu~Sun, Hao Tian, Hua Wu, and Haifeng Wang.
\newblock Ernie-vilg: Unified generative pre-training for bidirectional
  vision-language generation.
\newblock \emph{arXiv preprint arXiv:2112.15283}, 2021{\natexlab{b}}.

\bibitem[Zhou et~al.(2021)Zhou, Zhang, Chen, Li, Tensmeyer, Yu, Gu, Xu, and
  Sun]{zhou2021lafite}
Yufan Zhou, Ruiyi Zhang, Changyou Chen, Chunyuan Li, Chris Tensmeyer, Tong Yu,
  Jiuxiang Gu, Jinhui Xu, and Tong Sun.
\newblock Lafite: Towards language-free training for text-to-image generation.
\newblock \emph{arXiv preprint arXiv:2111.13792}, 2021.

\bibitem[Zhu et~al.(2019)Zhu, Pan, Chen, and Yang]{zhu2019dm}
Minfeng Zhu, Pingbo Pan, Wei Chen, and Yi~Yang.
\newblock Dm-gan: Dynamic memory generative adversarial networks for
  text-to-image synthesis.
\newblock In \emph{Proceedings of the IEEE/CVF conference on computer vision
  and pattern recognition}, pages 5802--5810, 2019.

\end{thebibliography}

\appendix
\section{Formulation of the Gaussian Diffusion Model}
We provide a detailed formulation of the Gaussian diffusion models from~\citet{ho2020denoising}.
For a data sample $x_0 \sim q(x_0)$, the Markovian noising process gradually adds noise to $x_0$ to produce noised samples $x_1$ through $x_T$. Each step of the Markovian noising process adds Gaussian noise according to some variance schedule $\beta_{t}$:
\begin{equation}
    q(x_t|x_{t-1}) = \mathcal{N}(x_t; \sqrt{1 - \beta_t} x_{t-1}, \beta_t \mathrm{I})
\label{eq_forward}
\end{equation}
For a given timestep $t$, $q(x_t|x_0)$ can be formulated as a Gaussian Distribution with $\alpha_t = 1 - \beta_t$ and $\bar{a}_t = \prod_{i=0}^{t} \alpha_i$:
\begin{align}
    &q(x_t|x_0) = \mathcal{N}(x_t; \sqrt{\bar{a}_t} x_0, (1 - \bar{a}_t) \mathrm{I}) \\
    &x_t = \sqrt{\bar{\alpha}_t} x_0 + \epsilon \sqrt{1 - \bar{\alpha}_t}, \epsilon \sim \mathcal{N}(0, \mathrm{I})
\label{eq_xt}
\end{align}

Thus, for a given $t \sim [1, T]$ and $x_t$, we can also approximately predict $\hat{x}_0$ by:
\begin{equation}
    \hat{x}_0 = \frac{1}{\sqrt{\bar{\alpha}_t}} (x_t - \sqrt{1 - \bar{\alpha}_t} \epsilon_t)
\label{eq_x0}
\end{equation}

Based on Bayes theorem, the posterior $q(x_{t-1}|x_t,x_0)$ is also a Gaussian distribution $\mathcal{N}(x_{t-1};\tilde{u}_t(x_t,x_0), \tilde{\beta}_t)$ with mean $\tilde{u}_t(x_t,x_0)$ and variance $\tilde{\beta}_t$ defined as follows:
\begin{align}
    \tilde{u}_t(x_t,x_0) &= \frac{\sqrt{\bar{\alpha}_{t-1}} \beta_t}{1 - \bar{\alpha}_t} x_0 + \frac{\sqrt{\alpha_t} (1 - \bar{\alpha}_{t-1})}{1 - \bar{\alpha}_t} x_t \\
    \tilde{\beta}_t &= \frac{1 - \bar{\alpha}_{t-1}}{1 - \bar{\alpha}_t} \beta_t
\label{eq_posterior}
\end{align}

In order to sample from the data distribution $q(x_0)$, diffusion models first sample from $q_(x_T)$ and then sample reverse steps $q(x_{t-1}|x_t)$ until reach $x_0$. The distribution $q(x_{t-1}|x_t)$ approaches a diagonal Gaussian distribution as $T \to \infty$ and correspondingly $\beta_t \to 0$, so it can be approximated by a neural network to predict a mean $u_{\theta}$ and a diagonal covariance matrix $\Sigma_{\theta}$:
\begin{equation}
    p_{\theta}(x_{t-1}|x_t) = \mathcal{N}(x_{t-1};u_{\theta}(x_t,t),\Sigma_{\theta}(x_t,t))
\label{eq_13}
\end{equation}
Instead of directing parameterize $u_{\theta}(x_t,t)$ by a neural network, \citet{ho2020denoising} found a simple formulation that predict the noise $\epsilon$ in Equation~\ref{eq_xt} by training a model $\epsilon_{\theta}(x_t,t)$. The simplified training objective is defined as in Equation~\ref{eq_loss}.
During sampling, we can derive $u_{\theta}(x_t,t)$ from $\epsilon_{\theta}(x_t,t)$:
\begin{equation}
    u_{\theta}(x_t,t) = \frac{1}{\sqrt{\alpha}_t} (x_t - \frac{1 - \alpha_t}{\sqrt{1 - \bar{\alpha}_t}} \epsilon_{\theta}(x_t,t))
\label{eq_14}
\end{equation}
The covariance matrix $\Sigma_{\theta}$ is usually fixed a a constant, choosing either $\beta_t \mathrm{I}$ or $\tilde{\beta}_t I$ which correspond to upper and lower bounds for the true reverse step variance.

\section{Comparison Examples}
We list several comparison examples between UPainting, Stable Diffusion and Disco Diffusion.

\begin{figure}
    \centering
    \setlength{\leftskip}{-90pt}
    \includegraphics[width=8in]{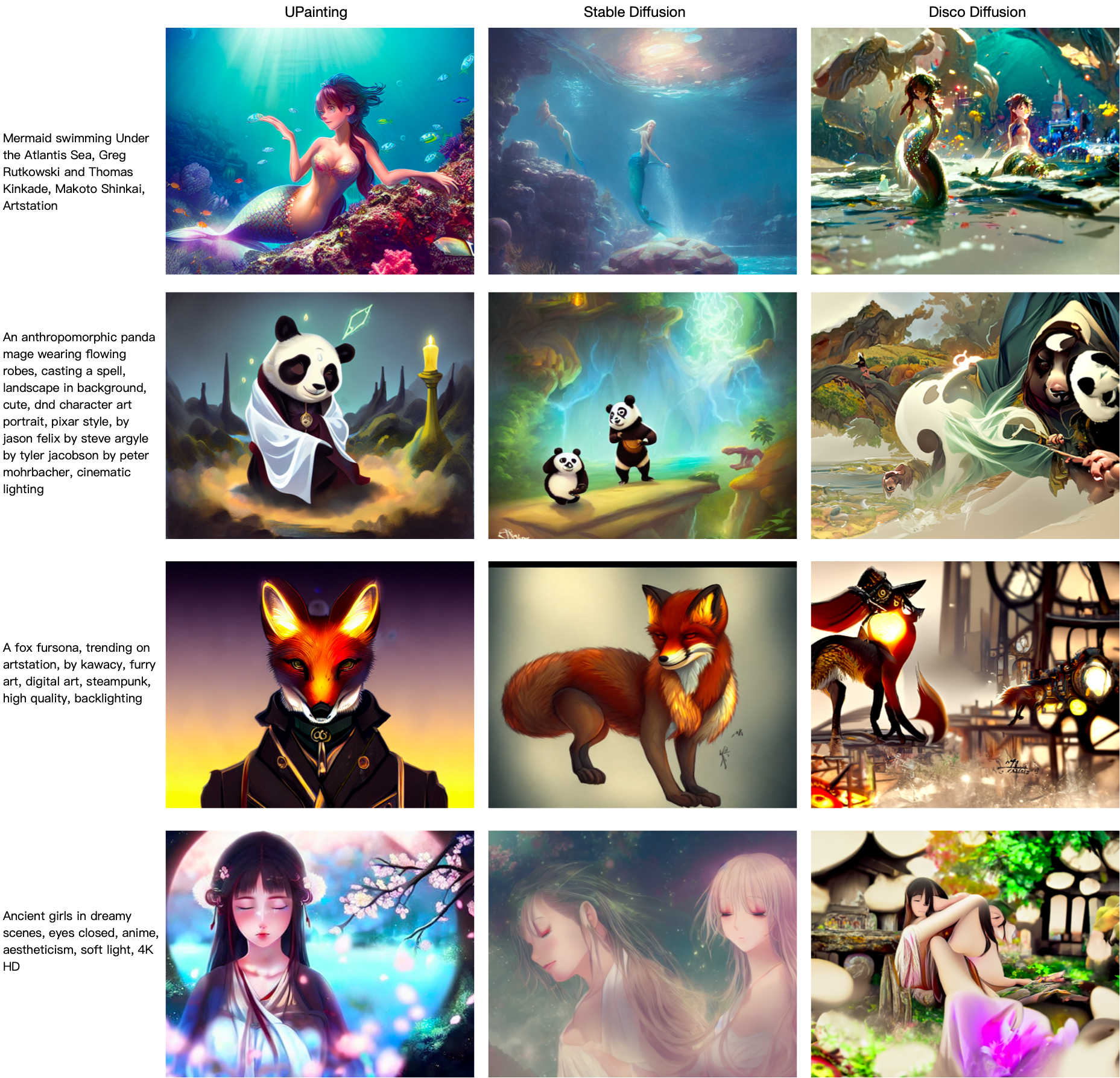}
    \caption{Selected samples of  simple-scene images.}
    \label{fig:comparison_small_scene}
\end{figure}

\begin{figure}
    \centering
    \setlength{\leftskip}{-90pt}
    \includegraphics[width=8in]{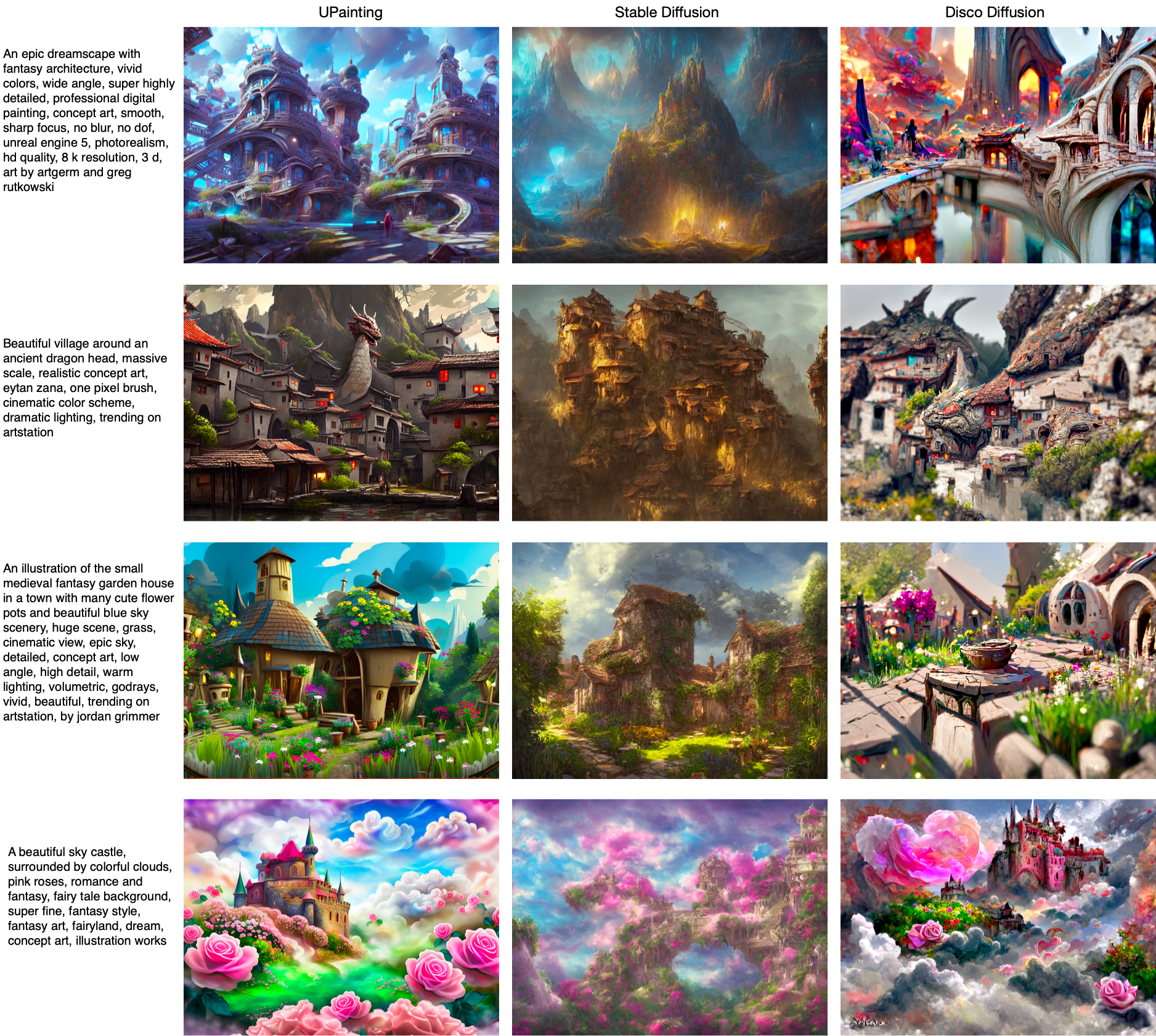}
    \caption{Selected samples of  complex-scene images.}
    \label{fig:comparison_complex_scene}
\end{figure}

\end{document}